\def\year{2022}\relax
\documentclass[letterpaper]{article} % DO NOT CHANGE THIS
\usepackage{aaai22}  % DO NOT CHANGE THIS
\usepackage{times}  % DO NOT CHANGE THIS
\usepackage{helvet}  % DO NOT CHANGE THIS
\usepackage{courier}  % DO NOT CHANGE THIS
\usepackage[hyphens]{url}  % DO NOT CHANGE THIS
\usepackage{graphicx} % DO NOT CHANGE THIS
\usepackage{natbib}  % DO NOT CHANGE THIS AND DO NOT ADD ANY OPTIONS TO IT
\usepackage{caption} % DO NOT CHANGE THIS AND DO NOT ADD ANY OPTIONS TO IT
\DeclareCaptionStyle{ruled}{labelfont=normalfont,labelsep=colon,strut=off} % DO NOT CHANGE THIS
\usepackage{algorithm}
\usepackage{algorithmic}

\usepackage[utf8x]{inputenc} 
\usepackage{multirow}
\usepackage{float} 
\usepackage{subfigure}

\usepackage{xcolor}
\usepackage{amsmath}
\usepackage{amssymb}
\usepackage{comment}
\usepackage{booktabs}
\usepackage[switch]{lineno}

\usepackage{makecell}
\newtheorem{theorem}{Theorem}  

\usepackage{newfloat}
\usepackage{listings}
\floatstyle{ruled}
\newfloat{listing}{tb}{lst}{}
\floatname{listing}{Listing}
\title{An Empirical Study and Analysis on Open-Set Semi-Supervised Learning}
\author{Huixiang Luo$^*$, Hao Cheng$^*$, Fanxu Meng, Yuting Gao,  Ke~Li, Mengdan Zhang, Xing Sun$^{\dagger}$\\ 
	Tencent Youtu Lab, Shanghai, China\\
	\tt \scriptsize \{luobosirobots,louischeng369,tristanli.sh,winfred.sun\}@gmail.com, \{rumimeng,yutinggao,davinazhang\}@tencent.com}

\begin{document}
	
	\maketitle

	%%%%%%%%% ABSTRACT
	\begin{abstract}
		Pseudo-labeling (PL) and Data Augmentation based Consistency Training (DACT) are two approaches widely used in Semi-Supervised Learning (SSL) methods. These methods exhibit great power in many machine learning tasks by utilizing unlabeled data for efficient training. But in a more realistic setting (termed as open-set SSL), where unlabeled dataset contains out-of-distribution (OOD) samples, the traditional SSL methods suffer severe performance degradation. Recent approaches mitigate the negative influence of OOD samples by filtering them out from the unlabeled data. However, it is not clear whether directly removing the OOD samples is the best choice. Furthermore, why PL and DACT could perform differently in open-set SSL remains a mystery. In this paper, we thoroughly analyze various SSL methods (PL and DACT) on open-set SSL and discuss pros and cons of these two approaches separately. Based on our analysis, we propose Style Disturbance to improve traditional SSL methods on open-set SSL and experimentally show our approach can achieve state-of-the-art results on various datasets by utilizing OOD samples properly. We believe our study can bring new insights for SSL research.
	\end{abstract}

	%%%%%%%%% BODY TEXT
	\section{Introduction}
\begin{figure}[t]
		\begin{center}
			\includegraphics[width=1.0\linewidth]{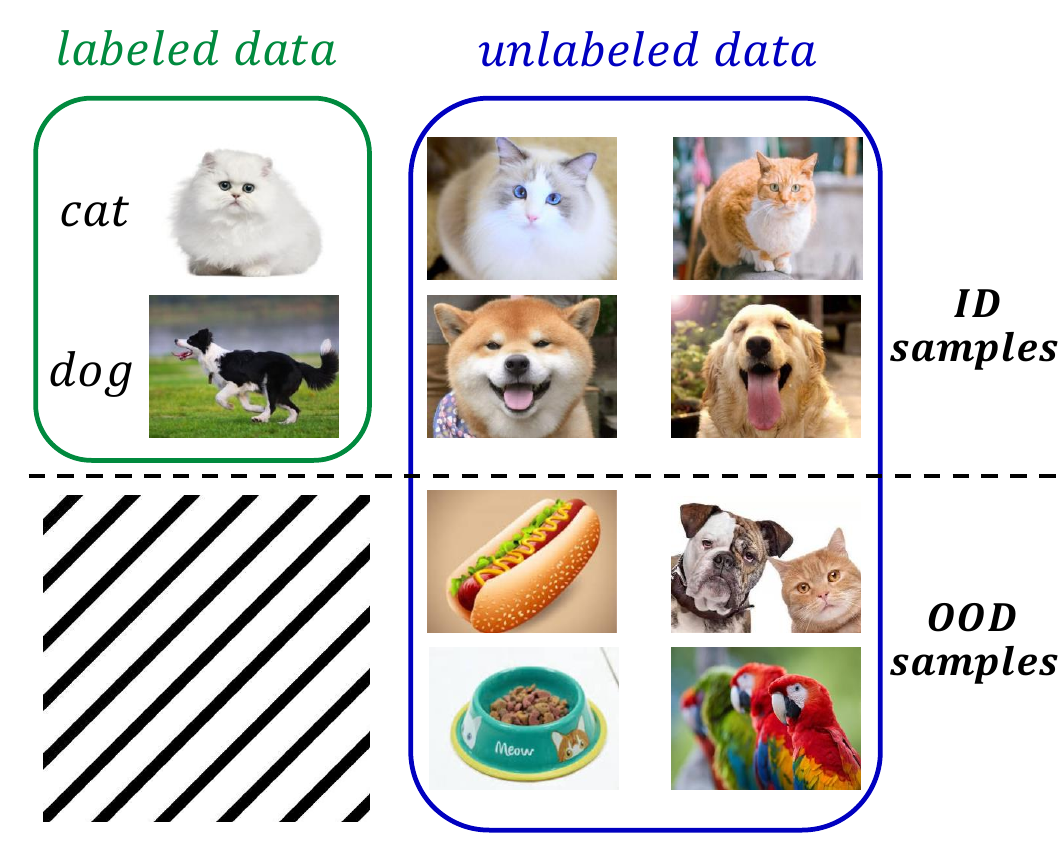}
		\end{center}
		\caption{In open-set semi-supervised learning, unlabeled dataset contains OOD samples that do not belong to any labeled single class.}
		\label{fig:open-set_SSL}
	\end{figure}

	\footnotetext[1] {In the author list, $^{\ast}$ denotes that authors contribute equally; $^{\dagger}$ denotes corresponding authors. The work is conducted while Huixiang Luo works as an internship at Tencent Youtu Lab. }
	The majority of SSL algorithms are designed assuming that both the labeled and the unlabeled dataset are drawn from the same distribution, which means they share the same classes and no outlier exists in the unlabeled dataset. 
	When it comes to a more realistic setting where the unlabeled dataset contains out-of-distribution (OOD) samples, the performance of many popular SSL algorithms is severely damaged \cite{oliver2018realistic, guo2020safe, yu2020multi, chen2020semi}. 
	This setting is firstly introduced by \cite{yu2020multi}, and is named as "Open-Set Semi-Supervised Learning" (open-set SSL, illustrated in Figure \ref{fig:open-set_SSL}). 
	To mitigate the negative influence of OOD samples, a straightforward approach \cite{chen2020semi, yu2020multi} is to roll back from the open-set SSL setting to the conventional SSL setting by detecting OOD samples at first and filtering them out later. 
	Presuming the ever-present harm to model performance caused by OOD samples, recent studies \cite{oliver2018realistic, yu2020multi, guo2020safe} weaken their impact on several levels during training, e.g. data-level, feature-level or loss-level. 
	
	However, it is unclear whether directly removing the OOD samples is the best choice for both PL and DACT based SSL methods. 
	In this paper, we deeply study the pros and cons of these two mainstream approaches and analyze the robustness when dealing with OOD samples. We find that: for PL-based methods, the OOD detection module is necessary; for DACT-based methods, this module is not a must. However, to make the method robust to OOD samples, diverse and carefully-searched augmentation strategies are required, and they are often costly to search out.
	
	Besides, the traditional performance upper bound from \cite{oliver2018realistic, yu2020multi, guo2020safe} on open-set SSL setting neglects the possible positive outcomes brought by OOD samples. 
	Inspired by the phenomena that the OOD dataset used for pretraining can help boost model performance on downstream tasks on ID datasets, we also exploit the potentialities of OOD samples. 
	In this paper, we argue that performance boundary can be extended by properly utilizing OOD samples: For DACT, the distribution gap between labeled and unlabeled data is reduced via Style Disturbance; for PL, OOD-style perturbations are added on features of ID samples in a similar way.
	%It is worth noting that our approach does not contradict with self-supervised learning since they improve the network in variant ways (see ablation study and discussion for details). 
	Experiments on several open-set SSL settings show the advantages of our methods over the others.
	The contributions of this paper can be summarized as follows:
	
	\begin{itemize}
		\item{
			We analyze two fundamental SSL methods: Pseudo Labeling (PL) and Data Augmentation based Consistency Training (DACT) on open-set SSL and observe that DACT is more robust than PL. However, requirement for diverse and carefully-searched data augmentation strategies for DACT can be costly. Also, we find there exists a close relationship between DACT on open-set SSL and self-supervised pre-training.
		} 
		
		\item{We propose Style Disturbance to make better use of OOD samples with either PL or DACT based SSL methods. 
			With Style Disturbance, our DACT based methods can surpass SOTA methods on several open-set SSL settings, because style-disturbed samples help reduce the distribution gap between labeled and unlabeled samples. 
			Our method also helps boost the performance of PL-based SSL methods by adding OOD-style perturbations on features of ID samples.
		} 
	\end{itemize}
	
	\section{Related Work}
	
	\subsection{Conventional semi-supervised learning.} 
	The goal of semi-supervised learning (SSL) methods is to leverage unlabeled data for performance improvement on labeled data. 
	%We focus on current SOTA methods for image classification tasks, so SSL techniques including graph-based methods \cite{kipf2016semi, jiang2019semi, lin2020shoestring, xu2020graph} and generative modeling \cite{kingma2014semi, ehsan2017infinite, odena2016semi} are not discussed here in detail. 
	In conventional semi-supervised learning, pseudo labeling (PL) and consistency regularization are two popular and fundamental methods of many recent SOTA algorithms \cite{sohn2020fixmatch, kuo2020featmatch, xie2020self, rizve2021defense}. 
	
	\noindent\textbf{Pseudo Labeling} takes the idea that model trained by labeled data can obtain artificial labels of unlabeled dataset by itself  \cite{lee2013pseudo, pham2020meta, li2020density}. 
	In a narrow sense, PL refers to using 'hard' pseudo-labels of samples that satisfy constraints of threshold $\eta$ (e.g. the maximal class probability $\ge \eta$ \cite{lee2013pseudo}). 
	The hard PL method is closely related to entropy minimization \cite{lee2013pseudo, grandvalet2005semi}, a common method forcing the model's prediction to be in high confidence on unlabeled samples. 
	In a broad sense, methods \cite{wang2020repetitive, pham2020meta} using 'soft' pseudo-label for supervision also belong to PL-based algorithms. 
	
	In addition to model prediction based PL methods, there also exist PL methods \cite{iscen2019label, zhu2002learning} based on label propagation. 
	
	\noindent\textbf{Consistency regularization}, another fundamental component in SSL algorithms, is usually implemented by enforcing the model output stable with perturbations during training. 
	%Consistency constraints are required in various ways, such as image-level \cite{berthelot2019mixmatch, berthelot2019remixmatch, sohn2020fixmatch, xie2020unsupervised, wang2019enaet}, model-level \cite{xie2020self, zhang2020wcp}, feature-level \cite{kuo2020featmatch}, distribution-level \cite{berthelot2019remixmatch} and temporal-level \cite{tarvainen2017mean, laine2016temporal, zhou2020time} consistency. 
	The most widely used consistency regularization is Data-Augmentation based Consistency Training (DACT). 
	MixMatch \cite{berthelot2019mixmatch} applies random horizontal flips and crops several times on a single image, and the average prediction is used for consistency training. 
	ReMixMatch \cite{berthelot2019remixmatch} requires the model's prediction of weakly and strongly augmented images to be consistent with each other. 
	UDA \cite{xie2020unsupervised} leverages CutOut \cite{devries2017improved} and RandAugment \cite{cubuk2020randaugment} to unlabeled samples. 
	FixMatch \cite{sohn2020fixmatch} follows UDA and ReMixMatch to adopt similar strategies as strong augmentation with samples filtered by the threshold.
	
	\subsection{Open-set semi-supervised learning}
	Open-set SSL is a more realistic setting mentioned in \cite{oliver2018realistic, yu2020multi}, where only 'dirty' unlabeled dataset (i.e., dataset has both In-Distribution (ID) and Out-Of-Distribution (OOD) samples) is available. 
	Algorithms \cite{xie2020unsupervised, pham2020meta, laine2016temporal} for conventional SSL settings tend to filter out OOD samples with a threshold in advance before using dirty unlabeled dataset. 
	%These offline 'hard-weight' methods (i.e., weight of OOD and ID samples is set to zeros and ones respectively) are later improved as 'soft-weight' and online filtering methods. 
	UASD \cite{chen2020semi} ensembles model predictions temporally, and the maximal probability of unlabeled samples are compared with a dynamic threshold to discard OOD samples online. 
	MTCF \cite{yu2020multi} takes the idea of Positive and Unlabeled (PU) learning; OOD detection and model training are performed simultaneously. 
	DS3L \cite{guo2020safe} designs an online 'soft-weight' framework (i.e. weight of samples $\in [0, 1]$ at loss level) formulated as a bi-level optimization problem.
	%Guided by meta-learning \cite{shu2019meta}, weight of OOD samples is reduced to zero softly on the loss level.
	RobustSSL \cite{zhao2020robust} combines meta-learning with Weighted Batch Normalization to reduce the influence of OOD samples at the feature level.
	Unlike previous methods handling OOD samples by weight reduction, our method could make better use of them by Style Disturbance.
	
	\subsection{Self-Supervised Learning} 
	In the field of computer vision, self-supervised learning aims to learn effective visual representations from unlabeled datasets. 
	The common way of self-supervised learning is a 2-step procedure: (1) pretrain a model on large-scale unlabeled datasets containing OOD samples; (2) finetune the pretrained model by labeled ID dataset. 
	Recent self-supervised learning algorithms \cite{he2020momentum, chen2020big, grill2020bootstrap, chen2020exploring} that apply multi-view consistency into pretraining achieve great performance on downstream ID tasks, and show a new way of leveraging OOD samples.
	%MoCo \cite{he2020momentum} is adopted as the self-supervised pretraining module of our method to verify its effectiveness in the Open-SSL setting. 
	We also compare DACT based semi-supervised learning and multi-view consistency-based self-supervised learning, analyze the differences and similarities to better understand their relationship.

	\section{Pseudo Labeling in open-set SSL}
	
	A general PL method first calculates the probability of all the unlabeled data belonging to each class, then carefully picks them out by some threshold(s), such as confidence threshold \cite{lee2013pseudo, rizve2021defense} and threshold of selected samples per class \cite{rizve2021defense}. 
	To test the robustness (i.e., whether the model performance drops after OOD samples are added into unlabeled dataset) of PL on OOD samples, we conduct experiments on several PL-based SSL methods in the open-set SSL setting.
	\subsection{Experiments} \label{sec_3_1}
	
	Pseudo-Label \cite{lee2013pseudo}, R2-D2 \cite{wang2020repetitive} and LabelProp \cite{iscen2019label}, 3 typical PL-based SSL methods are chosen for experiment on CIFAR-10 \cite{krizhevsky2009learning} (ID data), with 4 types of OOD data: Tiny ImageNet (TIN) \cite{deng2009imagenet}, LSUN \cite{yu2015lsun}, Gaussian noise (GN) and Uniform Noise (UN).
	
	\noindent\textbf{Result:} As is shown in Table \ref{table:PL_methods}, all methods suffer severe performance drop on CIFAR-10. Additional experiments conducted with Pseudo-Label on SVHN or TIN show similar results (refer to the supplementary material for detail). These experiments indicate that PL-based SSL methods are generally not robust on open-set SSL.
	
	\subsection{Problems of Pseudo Labeling}
	In this section, we analyze PL methods based on the most widely used threshold: confidence threshold. 
	The reason why PL cannot be robust when dealing with OOD samples are summarized as follows:
	
	\begin{itemize}
		\item The safe threshold changes when unlabeled ID samples are mixed with OOD samples.
		\item Unlabeled OOD samples may be labeled wrongly and intensively to several single classes, which causes more severe class-imbalance problems.
		\item OOD samples may disobey the preconditions of ID tasks differently. The precondition used in this paper is that each sample only belongs to a single class.
	\end{itemize}
	
\begin{figure}[t]
		\begin{center}
			\includegraphics[width=1.0\linewidth]{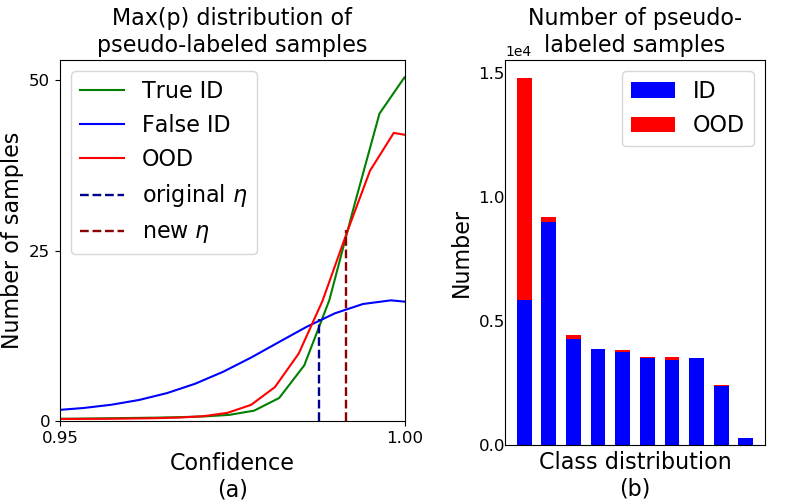}
		\end{center}
		\caption{Visualizing the changed threshold and class imbalance problems on the experiment of CIFAR-10 (ID data, 1k labeled, 54k unlabeled) $\&$ LSUN (OOD data, 10k) with Pseudo-Label. (a) Distribution of training samples’ maximum softmax probability: the safe threshold $\eta$ is getting closer to 1 after OOD samples are added in. (b) Number of pseudo-labeled samples per class: the top-2 classes containing most pseudo-labeled ID samples also get the majority of pseudo-labeled OOD samples.}
		\label{fig:thre_change}
		
	\end{figure}

	\noindent\textbf{Problem of changed threshold:} Denote $ID_{\eta}^{right}$ as the number of the selected samples whose pseudo labels are right according to threshold $\eta$, and $ID_{\eta}^{wrong}$ as number of the selected samples which are wrongly pseudo-labeled. The safe threshold $\eta_{safe}$ is defined as:
	\begin{equation}
		\eta_{safe}:= \{ \eta~|~ ID_{\eta}^{right} > ID_{\eta}^{wrong}   \}.
	\end{equation}
	This definition is referred to the setting of learning with noisy labels \cite{natarajan2013learning, cheng2020learning,zhu2021second,zhu2021clusterability}, which indicates that at least, the selected samples have meaningful information. 
	We draw a figure in Figure \ref{fig:thre_change} (a) showing the distribution of training samples' maximum softmax probability. 
	It can be seen that if unlabeled data containing OOD samples, $\eta_{safe}$ are required to be closer than 1, which means choosing a safe threshold gets harder. 
	Also since $ID_{\eta}^{right}$ decreases as $\eta$ gets larger, only increasing the $\eta$ may result in very few valid samples for further training.
	
	\begin{table*}
	\small{
		\begin{center}
			\centering
			\begin{tabular}{c|c|c|c|c|c|c|c|c}
				\hline
				\multirow{2}{*}{\makecell[c] {Labeled\\Samples}} &\multirow{2}{*}{\makecell[c] {Unlabeled\\Samples}}& \multirow{2}{*}{Method} &
				\multicolumn{5}{c|}{OOD dataset} & \multirow{2}{*}{\makecell[c] {Mean acc\\ change}} \\
				\cline{4-8}
				&&{} & Clean & LSUN & TIN & GN & UN & {} \\
					\hline\hline
				\multirow{3}{*}{1K}&\multirow{3}{*}{54K}&Pseudo-Label & 72.14 ± 1.40 & 67.04 ± 0.08          & 66.65 ± 1.63 & 69.09 ± 1.22 & 68.88 ± 2.05 & \color{red} $\downarrow$ 4.23 \\
				&&R2-D2 & 85.95 ± 0.05 & 80.25 ± 0.10          & 79.20 ± 0.06 & 81.80 ± 0.18 & 80.26 ± 0.11 & \color{red} $\downarrow$ 5.54 \\
				&&LabelProp  & 79.38 ± 0.06 & 75.30 ± 1.07 & 75.85 ± 0.89 & 77.80 ± 0.25 & 77.82 ± 0.11 & \color{red} $\downarrow$ 2.68 \\
				\hline
				\multirow{3}{*}{4K}&\multirow{3}{*}{51K}&Pseudo-Label & 84.84 ± 0.03 & 83.98 ± 0.11          & 83.15 ± 0.03 & 83.97 ± 0.23 & 84.07 ± 0.23 & \color{red} $\downarrow$ 1.04 \\
				&&R2-D2 & 93.97 ± 0.03 & 93.09 ± 0.06 & 92.75 ± 0.11 & 92.58 ± 0.04 & 93.57 ± 0.19 & \color{red} $\downarrow$ 0.97 \\
				&&LabelProp  & 87.67 ± 0.04 & 86.52 ± 0.12 & 85.95 ± 0.08 & 87.44 ± 0.19 & 87.54 ± 0.02 & \color{red} $\downarrow$ 0.80 \\
				\hline
			\end{tabular}
		\end{center}
		\caption{Performance of PL-based methods on CIFAR-10. 'Clean' means unlabeled dataset doesn't contain OOD samples. 5k samples are split from the original training data as validation set; the remaining samples are split into labeled and unlabeled dataset; number of labeled samples is chosen from \{1k, 4k\}. Each type of OOD dataset contain 10k samples. WRN-28-2 \cite{zagoruyko2016wide}, Shake-Shake \cite{gastaldi2017shake} and 13 Layer CNN \cite{iscen2019label} are used as backbone for Pseudo-Label \cite{lee2013pseudo}, R2-D2 \cite{wang2020repetitive} and LabelProp \cite{iscen2019label} respectively. We use the released codes for experiment and keep all hyper-parameters unchanged.}
		\label{table:PL_methods}
	}
	\end{table*}
	
	\noindent\textbf{Problem of class imbalance:} With OOD samples mixed in the unlabeled data, the inherent class-imbalance problem of PL might get worse.
	Recent study on PL based methods \cite{rizve2021defense} shows that: 
	pseudo-labeling with a limited number of labeled samples is difficult, especially for datasets with a small number of classes.
	For these datasets, the model tends to overfit early and be biased towards easy classes.
	%This is specifically true during the initial iterations.
	Study \cite{kim2020distribution} on the class-imbalanced SSL setting also finds that: 
	compared to true labels of labeled or unlabeled data, pseudo-labels are even more imbalanced.
	A biased model with imbalanced pseudo-labels results in a severe class-imbalanced SSL learning problem.
	%We perform experiments on CIFAR-10 (1k labeled samples, 54k unlabeled ID samples $\&$ 10k OOD samples from LSUN) with Pseudo-Labeling \cite{lee2013pseudo} to visualize this phenomenon.
	As is illustrated in Figure \ref{fig:thre_change} (b), the top-2 classes containing most pseudo-labeled ID samples also get the majority of pseudo-labeled OOD samples, thus the class distribution is more imbalanced.
	
	\noindent\textbf{Problem of precondition disobedience:} In a real-world scenario, samples may not satisfy the rule that each belongs to only one class, which is the precondition of multi-class single-label classification. 
	As is shown in Figure \ref{fig:open-set_SSL}, some OOD images may contain objects from several ID classes, while others may just contain objects from none of any ID classes. 
	These images may confuse the classifier differently and make the threshold harder to control. 
	Methods of multi-label learning \cite{durand2019learning} and complementary label learning \cite{feng2020learning} may be helpful to this problem.

	Due to the reasons above, OOD samples may cause large performance drop for PL method. 
	Therefore, using OOD detection module to discard OOD samples is a safe and direct way to mitigate their impact, and recent studies \cite{guo2020safe, chen2020semi, yu2020multi} on open-set SSL all adopt this strategy.

	\section{Data Augmentation based Consistency Training in Open-Set SSL}
	As a way of consistency regularization, DACT improves model performance by enforcing the model's output of one training sample stable to its perturbation, which is performed by Data Augmentation (DA) strategy on the image. 
	Recent study \cite{ghosh2021data} shows that: consistency-based SSL methods are more powerful when efficient DA strategies are used for exploiting the geometry of the data-manifold.
	Many SOTA methods on conventional SSL setting also coincide with this observation: 
	Costly DA strategies searched on the specific dataset are adopted, 
	e.g., UDA \cite{xie2020unsupervised} uses AutoAugment (AA) \cite{cubuk2018autoaugment}, 
	FixMatch \cite{sohn2020fixmatch} proposes CTAugment, 
	and FeatMatch \cite{kuo2020featmatch} applies RandAugment (RA) \cite{cubuk2020randaugment} to achieve great performance.
	We also follow their studies and 3 popular DA strategies, Fast AutoAugment (FA) \cite{lim2019fast}, AA and RA are tested on their robustness in the sections below.

		\begin{table}[t]
		\small{
			\begin{center}
				%\begin{subtable}{\textwidth}
				\centering
				\renewcommand\tabcolsep{1.5pt}
				\begin{tabular}{c|c|c|c|c}
					\hline
					Dataset&{DataAug}&OOD dataset&Acc&{Mean acc change}\\
					\hline\hline
					\multirow{9}{*}{CIFAR-10}&\multirow{3}{*}{AA}&Clean&88.87 ± 0.66&\multirow{3}{*}{\color{red} $\downarrow$ 1.61}\\
					&&LSUN&86.42 ± 0.21&\\
					&&GN&88.10 ± 0.17 &\\
					\cline{2-5}
					
					&\multirow{3}{*}{RA}&Clean&81.89 ± 0.43&\multirow{3}{*}{\color{red} $\downarrow$ 0.32}\\
					&&LSUN&81.48 ± 0.65&\\
					&&GN&81.77 ± 0.32 &\\
					\cline{2-5}
					
					&\multirow{3}{*}{FA}&Clean&88.29 ± 0.25&\multirow{3}{*}{\color{blue} $\uparrow$ 0.59}\\
					&&LSUN&88.60 ± 0.22&\\
					&&GN&89.15 ± 0.19 &\\
					\hline\hline
					
					\multirow{9}{*}{SVHN}&\multirow{3}{*}{AA}&Clean&95.54 ± 0.02&\multirow{3}{*}{\color{blue} $\uparrow$ 0.12}\\
					&&LSUN&95.49 ± 0.30&\\
					&&GN&95.82 ± 0.05  &\\
					\cline{2-5}
					
					&\multirow{3}{*}{RA}&Clean&92.07 ± 0.34&\multirow{3}{*}{\color{red} $\downarrow$ 0.25}\\
					&&LSUN&91.40 ± 0.42&\\
					&&GN&92.23 ± 0.35 &\\
					\cline{2-5}
					
					&\multirow{3}{*}{FA}&Clean&94.85 ± 0.24&\multirow{3}{*}{\color{blue} $\uparrow$ 0.31}\\
					&&LSUN&95.19 ± 0.29&\\
					&&GN&95.13 ± 0.51 &\\
					\hline\hline
					
					\multirow{6}{*}{TIN-20}&\multirow{2}{*}{AA}&Clean&54.55 ± 0.05&\multirow{2}{*}{\color{red} $\downarrow$ 3.15}\\
					&&TIN-180&51.40 ± 0.50&\\
					\cline{2-5}
					&\multirow{2}{*}{RA}&Clean&51.65 ± 0.65&\multirow{2}{*}{\color{blue} $\uparrow$ 1.70}\\
					&&TIN-180&53.35 ± 0.35&\\
					\cline{2-5}
					&\multirow{2}{*}{FA}&Clean&53.15 ± 0.05&\multirow{2}{*}{\color{blue} $\uparrow$ 0.75}\\
					&&TIN-180&53.90 ± 0.10&\\
					\hline
					
				\end{tabular}
				\caption{Performance of different DA strategies on CIFAR-10 (1k labeled, 55k unlabeled), SVHN (1k labeled, 77k unlabeled) and Tiny-ImagneNet (1k labeled, 99k unlabeled, only the last 20 classes are selected as ID classes).}
				\label{table:DA performance}
			\end{center}
		
		}
	\end{table}
	
	\subsection{Experiments}\label{sec_4_1}
	
	The DACT module of UDA \cite{xie2020unsupervised} is selected for robustness assessment on  CIFAR-10, SVHN and TIN, with DA strategies of AA, RA and FA separately.
	%\subsection{Result and analysis on the robustness of DACT}
	
	\noindent\textbf{Result:}
	Table \ref{table:DA performance} show how DACT performs in various DA strategies.
	Surprisingly, OOD samples may not always harm the model performance; FA could make use of OOD samples to slightly boost performance. These phenomena raise two questions to be discussed below:
	
	\subsubsection{How DACT based SSL works in open-set SSL?}\label{sec_4_2_1}
	Observing the similar phenomena of using OOD samples for performance improvement in self-supervised learning \cite{he2020momentum, caron2020unsupervised} $\&$ transfer learning \cite{kolesnikov2019big, mahajan2018exploring}, we turn to these fields for an explanation.
	We compare DACT on open-set SSL with multi-view consistency regularization on self-supervised learning (one possible way of transfer learning for using OOD samples) in two dimensions: the learning objective and experimental phenomena.
	
	\noindent\textbf{Dimension of learning objective:} 
	Denote $D_{ID}$ as ID data, and $D_{OOD}$ as unlabeled OOD samples, the learning objective of DACT on open-set SSL can be written as:
	
	\begin{equation}\label{open_ssl_eq}
		\mathbf{w}^{*} = \arg\min_{\mathbf{w}} L_{1}(\mathbf w, D_{ID}; \mathbf{w}_{ini}) 
		+ L_{2}(\mathbf w, D_{OOD}; \mathbf{w}_{ini}),
	\end{equation}
	where $\mathbf{w}_{ini}$ is the random initialized weight, $L_{2}$ is the DA-based consistency loss on OOD samples. 
	
	For multi-view consistency-based self-supervised learning, we first train unlabeled samples (OOD samples) via multi-view consistency loss, which is similar to $L_2$ on OOD samples. 
	Then the obtained weight is used to initialize the network for downstream tasks (ID sampels). 
	Thus mathematically:
	
	\begin{equation}\label{ssl_eq}
		\begin{aligned}
			\mathbf{w}^{*} = &\arg\min_{\mathbf{w}} L_{1}(\mathbf w, D_{ID}; \mathbf{w}^{*}_{L_{2}})\\
			{\rm ~s.t.} \mathbf{w}^{*}_{L_{2}} = &\arg\min_{\mathbf{w}} L_{2}(\mathbf w, D_{OOD}; \mathbf{w}_{ini}),
		\end{aligned}
	\end{equation}
	
	\begin{comment}
	Normally, solving \eqref{ssl_eq} involves two-stage training steps, as is seen in most self-supervised learning literatures \cite{he2020momentum, caron2020unsupervised, grill2020bootstrap, chen2020exploring, chen2020simple}. While solving \eqref{open_ssl_eq} is one-stage training process.
	By linking \eqref{ssl_eq} to \eqref{open_ssl_eq}, We have the following theorem:
	\begin{theorem}\label{theorem}
		Suppose both \eqref{open_ssl_eq} and \eqref{ssl_eq} converge to the global minimum. 
		Then \eqref{open_ssl_eq} equals to \eqref{ssl_eq} if the solution set of $L_{1}$ is the subset of the solution set of $L_{2}$.
	\end{theorem}
	
	Proof is left to the supplementary material. The condition in Theorem \ref{theorem} implies that only optimizing $L_{1}$ in the ID domain can also achieve robustness/generalization in the OOD domain. It is true that we expect our trained classifier can also generalize to other domains without re-training, but in general, existing studies show that generalization is not easy to achieve with limited ID samples (referring to Domain Generalization \cite{li2017deeper}). Besides, sometimes the network trained by ID samples is hard to converge and shows poor performance. Thus, pre-training on OOD samples may help the network converge faster and exhibit more robustness. 
	
	From the analysis, since the condition in Theorem \ref{theorem} cannot hold strictly, \eqref{open_ssl_eq} is an approximation of \eqref{ssl_eq}. But very similar experimental phenomena can be observed.
	\end{comment}
	
	One can see the difference between Equation \eqref{open_ssl_eq} and Equation \eqref{ssl_eq} lies in the optimization.  In Equation \eqref{ssl_eq}, $\mathbf{w}^{*}_{L_{2}}$ serves as a better initialization to help network optimize on ID samples while in Equation \eqref{open_ssl_eq}, we simultaneously optimize $L_{1}$ and $L_{2}$ in one loss function. For traditional open-set ssl settings, the number of ID samples is very small, thus $L_{2}$ can also help network optimize better without over-fitting to very few ID samples. 
	This is very different with traditional approaches \cite{yu2020multi} \cite{guo2020safe} that view OOD samples as negative effects to model performance, we show that by utilizing OOD samples properly, they can actually improve model performance and robustness. We further provide more experimental evidence to show the similarity between self-supervised pre-training (Equation \eqref{ssl_eq}) and our DACT approach (\eqref{open_ssl_eq}):

	\noindent\textbf{Dimension of experimental phenomena:} 
	We list similar experimental phenomena below:
	\begin{itemize}
		\item DA strategies greatly influence model performance.
		In self-supervised pretraining, either DA strategies made up of few DAs, or DAs not fit to (downstream) ID tasks result in bad performance \cite{tian2020makes, xiao2020should, chen2020simple, zbontar2021barlow};
		DA strategies more suitable to downstream tasks lead to better performance \cite{wang2020stronger, caron2020unsupervised, chen2020simple}. 
		In our experiment, we also observe that
		better DA policies searched for specific datasets show more performance improvement with OOD samples. (e.g., FA on Table \ref{table:DA performance}).
		
		\item 
		The distribution gap between the source domain (OOD samples) and target domain (ID samples) is a key factor that may degrade model performance.
		In transfer learning, the phenomena of negative transfer \cite{wang2019characterizing, raghu2019transfusion, ge2014handling, gui2018negative} are usually caused by large distribution gap, which can be calculated quantitatively by metrics such as Maximum Mean Discrepancy (MMD) \cite{gretton2007kernel, long2013transfer}.
		The same metric could be applied for open-set SSL as well, and we observe the similar phenomena that a larger gap between ID \& OOD data would make DACT less robust (e.g., LSUN v.s. GN on Table \ref{table:DA performance}) on OOD samples (refer to the supplementary material for detail).
		
	\end{itemize}
	
	According to the similarities (learning objective \& experimental phenomena) between DACT on open-set SSL and multi-view consistency regularization on self-supervised learning, the explanation of why self-supervised learning works can also apply to DACT on open-set SSL. There are many works explaining why self-supervised leaning works \cite{tsai2020demystifying, asano2019critical, zhao2020makes}. For example, rephrasing the opinion from \cite{tsai2020demystifying} in the context of open-set ssl: OOD samples can push model to extract task-relevant information and discard task-irrelevant information, helping model generalize well.

	\subsubsection{Why some DA are more robust to OOD samples?}
	In this section, we quantitatively explore the performance variation in terms of DA strategies.
	
	Recent studies \cite{wu2020generalization, raphael2020tradeoff} on label-preserving DA strategies try to understand the underlying mechanism behind their effectiveness. 
	Label-preserving transformations are well studied in \cite{wu2020generalization}, and their positive effect on model performance attributes to the amount of useful information provided by DA strategies. 
	They also apply the loss as a metric for picking out the most useful DAs: the higher the loss, the more information can be provided.
	In \cite{raphael2020tradeoff}, a similar metric $\mathcal{D}[a; m; D_{train}]$ named after '\textbf{diversity}', is used to evaluate the performance of DAs:
	\begin{equation}\label{diversity_eq}
		\mathcal{D}[a; m; D_{train}] := \mathbb{E}_{D^{'}_{train}}[L_{train}]
	\end{equation}
	where $a$ is a DA strategy, $D^{'}_{train}$ is the augmented training data resulting from $a$, and $L_{train}$ is the training loss for model $m$ trained on $D^{'}_{train}$. 
	
	We also adopt the metric to evaluate the usefulness of DA strategies. 
	We compare the metric between the generally worst strategy AA, and the best DA strategies among \{RA, FA\} on CIFAR-10, SVHN and TIN datasets.  
	For simplicity, we calculate the expectation of training loss by sampling its value every 400 epochs.
	
	\begin{table}[t]
	\small{
		\begin{center}
			\centering
			\renewcommand\tabcolsep{1.5pt}
			\begin{tabular}{c|c|c|c}
				\hline
				\multirow{2}{*}{ID Dataset}&\multirow{2}{*}{OOD dataset}&\multicolumn{2}{c}{DataAug}\\
				\cline{3-4}
				&&AA&FA\\
				\hline\hline
				\multirow{3}{*}{CIFAR-10}&Clean&1.17&1.27\\
				&LSUN&1.19&1.25\\
				&GN&1.06&1.13\\
				\hline\hline
				
				\multirow{3}{*}{SVHN}&Clean&0.93&1.05\\
				&LSUN&0.87&0.98\\
				&GN&0.85&0.96\\
				\hline\hline
				
				\multirow{2}{*}{TIN-20}&Clean&1.12&1.23\\
				&TIN-180&1.63&1.17\\
				\hline
			\end{tabular}
			\caption{Loss of different DA strategies on CIFAR-10 with 1k labeled and 54k unlabeled samples. }
			\label{table:DA performance_re}
		\end{center}
	}
\end{table}

\begin{comment}
		\begin{table}
		\small{
			\begin{center}
				%\begin{subtable}{\textwidth}
				\centering
				\renewcommand\tabcolsep{1.5pt}
				\begin{tabular}{c|c|c|c|c}
					\hline
					Dataset&{DataAug}&Clean&OOD Dataset&Loss\\
					\hline\hline
					\multirow{4}{*}{CIFAR-10}
					&\multirow{2}{*}{AA}&\multirow{2}{*}{1.17}&LSUN&1.19\\
					&&&GN&1.06\\
					\cline{2-5}
					
					&\multirow{2}{*}{FA}&\multirow{2}{*}{1.27}&LSUN&1.25\\
					&&&GN&1.13 \\

					\hline\hline
					\multirow{4}{*}{SVHN}
					&\multirow{2}{*}{AA}&\multirow{2}{*}{0.93}&LSUN&0.87\\
					&&&GN&0.85\\
					\cline{2-5}
					
					&\multirow{2}{*}{FA}&\multirow{2}{*}{1.05}&LSUN&0.98 \\
					&&&GN&0.96\\
					
					\hline\hline
					\multirow{2}{*}{TIN-20}
					&AA & 1.12&\multirow{2}{*}{TIN-180}& 1.23\\
					\cline{2-3}
					\cline{5-5}
					&FA & 1.63&& 1.17\\
					\hline

				\end{tabular}
				\caption{Performance of different DA strategies on CIFAR-10 with 1k labeled and 54k unlabeled samples. We follow the same setting of Sec. \ref{sec_3_1}.}
				\label{DA C10}
			\end{center}
			\label{table:DA performance_re}
		}
	\end{table}
	\end{comment}
	
	As is shown in Table \ref{table:DA performance_re}, better DA strategies own higher losses than the worst ones, so the experiment results are in line with the theories: \textbf{more diverse DA strategies could bring more useful information to model from OOD samples to improve performance}.
	%Sharing the same DA search space, we can also directly count the combinatorial number of all possible DA pairs as a rough estimation of diversity.
	
	\subsection{DACT v.s. PL}
	Our experiments show that DACT-based SSL methods are far more robust than PL-based methods on open-set SSL.
	For DACT, a properly chosen DA strategy is the key factor of robustness.
	However, piles of work are required to search for a robust strategy:
	\begin{itemize}
		\item
		{
			For a specific dataset, suitable DA methods with proper intensity are required to be selected, in order to construct a good search space.
		} 
		
		\item
		{
			Plenty of time and computing resources are necessary to search for a diverse strategy with trial and error.
		} 
	\end{itemize} 
	For lack of abundant domain-specific DA methods in many domains (e.g., videos and medical images \cite{shorten2019survey, yun2020videomix}), DACT-based methods might be a costly choice.
	
	Despite the fact that PL-based methods are not robust to OOD samples, they are domain-agnostic \cite{rizve2021defense}, and usually orthogonal to DACT \cite{rizve2021defense, cascante2020curriculum}, which means they can also benefit from diverse DA strategies.
	In other words, PL-based SSL methods are more general than DACT-based ones, as long as we have a good OOD detection module.
	
	\section{Style disturbance with OOD Samples}
	
	Both the success of self-supervised learning \cite{caron2020unsupervised, he2020momentum} and DACT experiments in Sec. \ref{sec_4_1} show the possibility of further improving model performance with OOD samples. 
	However, using OOD samples with these methods directly is not safe.
	As one way trying to transfer knowledge from OOD datasets, self-supervised learning methods may result in the problem of negative transfer \cite{xiao2020should, wang2019characterizing, raghu2019transfusion} because of the distribution gap between ID $\&$ OOD samples. 
	%The negative transfer phenomena are especially noticeable when the distribution gap is large since little useful knowledge can be transferred from OOD samples.

	To avoid negative transfer, one safe and efficient way is to add label-invariant perturbations with OOD samples.
	Neural Style Transfer (NST) algorithms could be adopted as label-preserving DA \cite{zheng2019stada, yin2016content, chen2016towards, jackson2019style} to use OOD samples.
	However, applying NST algorithms directly as DA could also hurt model performance \cite{jackson2019style}, since style information correlates strongly with class label in some datasets and removing the correlation by style transfer would lead to a performance drop.
	
	Due to the reasons above, we borrow the idea of AdaIN \cite{huang2017arbitrary} and mixup \cite{zhang2017mixup}, and 'disturb' styles of ID samples with OOD ones instead of wild style transfer.
	The core idea of our method is 'style disturbance' (i.e., the ratio of OOD style is far less than that of ID style when mixing them up), and it could be used in either image level (for DACT) or feature level (for PL).
	
	\subsection{Formulation}
	
	\subsubsection{Style disturbance for DACT}
	Mathematically, each sample $x$ could be split into two parts: $x = (x_{content}, x_{style})$, where $x_{content}$ preserves the information of class label and $x_{style}$ brings variance to the dataset. 
	AdaIN \cite{huang2017arbitrary}, a real-time arbitrary style transfer method is chosen for style disturbance. 
	The style-disturbed image $x_{i}^{sd}$ is generated with content of ID image $x_{i}^{ID}$ and style of OOD image $x_i^{OOD}$. 
	To avoid negative effect of artifacts caused by style transfer, $x_{i}^{sd}$ is further linearly interpolated with $x_{i}^{ID}$: 
	\begin{equation}
		x_{i}^{sd} = \beta AdaIN(x_i^{OOD}, x_{i}^{ID}, \omega) + (1-\beta) x_{i}^{ID}
	\end{equation}
	where both $\beta$ and $\omega \in [0, 1]$ control the similarity of $x_{i}^{sd}$ to $x_{i}^{ID}$, $AdaIN$ is the style-transfer network. 
	The style-disturbed dataset $D^{SD}$ would be used in the same way as original unlabeled dataset $D_U$ to keep consistency of model predictions with KL-divergence; 
	a standard cross-entropy loss is used for labeled dataset $D_L$. 
	The overall loss function is composed of the UDA loss $\mathcal{L}_{UDA}$ and loss of style-disturbed samples $\mathcal{L}_{SD}$. 
	It can be written as:
	\begin{equation}
		\mathcal{L}_{total} = \mathcal{L}_{UDA} + \mathcal{L}_{SD} = \mathcal{L}_{sup} + \mathcal{L}_{unsup} + \mathcal{L}_{SD}
	\end{equation}
	\begin{equation}
		\mathcal{L}_{sup} = \sum\limits_{x_l, y_l \in D_L}CE(y_l, p_{\phi}(y|x_l))
	\end{equation}
	\begin{equation}
		\mathcal{L}_{unsup} = \lambda_{u} \sum\limits_{x_u \in D_U} KL(p_{\phi}^{t}(y|x_u), p_{\phi}(y|Aug(x_u)))
	\end{equation}
	\begin{equation}
		\mathcal{L}_{SD} = \sum\limits_{x_{i}^{sd} \in D^{SD}}KL(p_{\phi}(y|x_{i}^{sd}), p_{\phi}(y|Aug(x_{i}^{sd})))    
	\end{equation}
	where $\lambda_u$ is the weighting coefficient of consistency loss, $\phi$ is the model and $Aug$ is Fast AutoAugment \cite{lim2019fast}.
	
	\begin{comment}
	\begin{table*}[htbp]
		\begin{center}
			\centering
			\begin{tabular}{c|c|c|c|c|c|c}
				\hline
				\multirow{2}{*}{Dataset}&
				\multicolumn{5}{c|}{OOD dataset} & \multirow{2}{*}{\makecell[c]{Mean acc\\ change}} \\
				\cline{2-6}
				& Clean & LSUN & TIN & GN & UN & {} \\
				\hline\hline
				CIFAR-10&55.50 ± 0.73&57.14 ± 0.74&56.90 ± 0.70&56.53 ± 0.29&56.60 ± 0.46 &\color{blue} $\uparrow$ 1.29 \\
				\hline
				TIN 20&46.97 ± 0.45&---&47.90 ± 1.07&---&---&\color{blue} $\uparrow$ 0.93\\
				\hline
			\end{tabular}\label{MM-UDA-DS3L-1k}
		\caption{Performance of Pseudo-Labeling with MixStyle on CIFAR-10 and TIN 20}
		\label{table:PL and MixStyle}
		\end{center}
	\end{table*}
	\end{comment}
	
	\begin{table*}[t]
		\begin{center}
			\centering
			\begin{tabular}{c|c|c|c|c|c|c|c|c}
				\hline
				\multirow{2}{*}{\makecell[c]{Labeled\\Samples}}&\multirow{2}{*}{\makecell[c]{Unlabled\\Samples}}&\multirow{2}{*}{Method} &
				\multicolumn{5}{c|}{OOD dataset} & \multirow{2}{*}{\makecell[c]{Mean acc\\ change}} \\
				\cline{4-8}
				&&{} & Clean & LSUN & TIN & GN & UN & {} \\
				\hline\hline
				
				\multirow{3}{*}{1K}&\multirow{3}{*}{54K}&DS3L & 67.79 ± 0.27 & 69.74 ± 0.08          & 70.10 ± 0.47 & 62.86 ± 0.67 & 62.89 ± 1.65 & \color{red} $\downarrow$ 1.39 \\
				&&MTCF & 90.67 ± 0.29 & 90.19 ± 0.47          & 89.85 ± 0.11 & 89.87 ± 0.08 & 89.80 ± 0.26 & \color{red} $\downarrow$ 0.74 \\
				&&Ours                    & 88.29 ± 0.25 & \textbf{91.30 ± 0.36} & \textbf{91.10 ± 0.65} & \textbf{92.33 ± 0.59} & \textbf{91.82 ± 0.04} & \color{blue} $\uparrow$ 3.35 \\
				\hline\hline
				\multirow{3}{*}{4K}&\multirow{3}{*}{51K}&DS3L  & 83.23 ± 0.07 & 82.89 ± 0.69 & 82.58 ± 0.14 & 80.44 ± 0.01 & 80.59 ± 0.03 & \color{red} $\downarrow$ 1.61 \\
				&&MTCF & 93.30 ± 0.10 & 92.91 ± 0.03 & 93.03 ± 0.05 & 92.83 ± 0.04 & 92.53 ± 0.08 & \color{red} $\downarrow$ 0.48 \\
				&&Ours                    & 93.36 ± 0.40 & \textbf{94.27 ± 0.21} & \textbf{93.84 ± 0.10} & \textbf{94.52 ± 0.07} & \textbf{94.50 ± 0.13} & \color{blue} $\uparrow$ 0.92 \\
				\hline
			\end{tabular}
			\caption{Experiments of CIFAR-10 with (1k labeled, 54k unlabeled) and (4k labeled, 51k unlabeled) samples}
			\label{table:MixMatch-vs-UDA-vs-DS3L-c10}
		\end{center}
		
	\end{table*}

	\subsubsection{Style disturbance for PL}
	For PL-based methods, we implement feature-level style disturbance on both $D_U^{ID}$ and $D_L^{ID}$ based on MixStyle \cite{zhou2021domain}, a Domain Generalization method that also combines AdaIN \cite{huang2017arbitrary} with mixup \cite{zhang2017mixup}.
	Given three batches of input: labeled batch $B_L$, unlabeled ID batch $B_U^{ID}$ and OOD batch $B_U^{OOD}$, the mixed feature statistics are computed as follows:
	\begin{equation}
		\sigma^{sd} = \rho \sigma(x^{id}) + (1-\rho)\sigma(x^{ood})
	\end{equation}
	\begin{equation}
		\mu^{sd} = \rho \mu(x^{id}) + (1-\rho)\mu(x^{ood})
	\end{equation}
	where $x^{id}$ comes from $B_L$ and $B_U^{ID}$, $x^{ood}$ comes from $B_U^{OOD}$, $\rho \in \mathbb{R}^B$ are instance-wise weights sampled from the Beta distribution Beta(9, 1),  $\mu^{sd}$ and $\sigma^{sd} $ are mean and standard deviation computed across the spatial dimension within each channel of each tensor. 
	Finally, the style-disturbed feature $x^{sd}$ is computed by applying the mixed feature statistics to style-normalized $x^{id}$:
	\begin{equation}
		x^{sd} = \sigma^{sd} \frac{x^{id} - \mu(x^{id})}{\sigma(x^{id})} + \mu^{sd}
	\end{equation}
	In practice, whether the MixStyle module is activated or not in the forward pass is according to a probability of 0.5; no MixStyle is applied at test time; gradients are blocked in the computational graph of $\mu$ and $\sigma$.
	After we get style-disturbed dataset $D^{SD}$, the original labels / pseudo-labels on $D_L$ and $D_U^{ID}$ are applied to $D_L^{SD}$ and $D_U^{SD}$ respectively
	%as supervision for PL-based methods
	, and the corresponding loss is $L_{SD}$.
	The total loss is: 
	\begin{equation}
		\begin{aligned}
			\mathcal{L}_{total} = \mathcal{L}_{PL} &+ \lambda_u \mathcal{L}_{SD} = \mathcal{L}_{sup}(D_L) + \mathcal{L}_{unsup}(D_U^{ID}) \\ 
			&+ \lambda_u (\mathcal{L}_{sup}(D_L^{SD}) + \mathcal{L}_{unsup}(D_U^{SD}) )
		\end{aligned}
	\end{equation}

		\begin{table}[t]
		\begin{center}
			\centering
			\renewcommand\arraystretch{0.9}
			\begin{tabular}{c|c|c|c}
				\hline
				ID dataset              & OOD dataset    & Accuracy  & \makecell[c]{Mean acc\\ change}  \\ \hline
				\multirow{5}{*}{CIFAR-10} & Clean   &  55.50 ± 0.73 & \multirow{5}{*}{\color{blue} $\uparrow$ 1.29} \\ 
				{}              & LSUN  & 57.14 ± 0.74        & {} \\ 
				{}              & TIN    & 56.90 ± 0.70   & {} \\ 
				{}              & GN       & 56.53 ± 0.29    & {} \\ 
				{}              & UN        & 56.60 ± 0.46   & {} \\ \hline
				\multirow{2}{*}{TIN 20} & Clean & 46.97 ± 0.45 & \multirow{2}{*}{\color{blue} $\uparrow$ 0.93} \\ 
				{}              & TIN 180 & 47.90 ± 1.07 & {} \\ \hline
			\end{tabular}
		\end{center}
		\caption{Performance of Pseudo-Labeling with MixStyle on CIFAR-10 and TIN 20}
		\label{table:PL and MixStyle}
	\end{table}

	\subsection{Experiments}\label{sec_5_1}
	Because our method of style disturbance is based on both ID $\&$ OOD data, OOD samples are required to be sorted out. 
	Since we just want to verify our hypothesis that OOD samples can be useful with style disturbance, the design of OOD detection module is not that important for our method.
	For DACT, we directly use a simplified detection module of \cite{yu2020multi} (refer to supplementary material for details);
	for PL, we suppose that all OOD samples are picked out perfectly.
	
	\subsubsection{Experiment setting for DACT}
	\noindent\textbf{Dataset:} Following conventional settings on OpenSet SSL  \cite{yu2020multi, oliver2018realistic, xie2020unsupervised, berthelot2019mixmatch} (the same setting as Sec.\ref{sec_4_1}), we use CIFAR-10 and TIN to perform experiments. 
	
	\noindent\textbf{Backbone:} We employ WRN-28-2 \cite{zagoruyko2016wide} and ResNet-50 \cite{he2016deep} for CIFAR-10 and TIN respectively. 
	
	\noindent\textbf{DACT module:} 
	This module is a submodule of UDA \cite{xie2020unsupervised}. 
	The model is trained for 1,600 epochs by default. 
	For CIFAR-10, batch sizes for labeled $\&$ unlabeled dataset are 36 and 960, softmax temperature is 0.8, $\lambda_{u}$ is 5, learning rate is 1e-4, learning rate schedule is cosine learning rate decay schedule with learning rate warming up for 120 epochs. 
	We use an SGD optimizer with nesterov momentum, momentum hyper-parameter is set to 0.9. 
	For TIN, we subtly modify the hyper-parameters above: unlabeled batch size is reduced to 64, softmax temperature is 1.
	
	\noindent\textbf{Style disturbance module.} 
	This module is based on AdaIN \cite{huang2017arbitrary}. 
	The style transfer network is an encoder-decoder framework, the encoder is a pretrained VGG-19 \cite{simonyan2014very} network, and the decoder in a reversed architecture is trained with style dataset $D_U$ and content dataset $D_L$. 
	We train the model with default hyper-parameters. 
	All images are resized to 224 x 224 in both model training and image generation procedures. 
	For CIFAR-10, generated images are further resized to 32 x 32. 
	The interpolation hyper-parameters $\beta$ and $\omega$ are randomly chosen from [0, 0.5]. 
	Unlabeled OOD samples and labeled samples are treated as style and content dataset respectively. 
	
	\subsubsection{Experiment setting for PL}
	\noindent\textbf{Dataset:} we follow the same setting as Sec. \ref{sec_3_1}: CIFAR-10 and TIN are used for experiments (1k labeled samples). 
	
	\noindent\textbf{Backbone:} We directly use ResNet-18 \cite{he2016deep} as the backbone for simplicity, since MixStyle is implemented on it according to the source code. Unlike the original use of MixStyle in \cite{zhou2021domain} where it is applied after several residual blocks, we only use it after the first block.
	
	\noindent\textbf{PL module:} This module is the same as Pseudo-Label \cite{lee2013pseudo}. We use the default hyper-parameters of the original implementation, and $\lambda_{u}$ is 0.5.

	\begin{comment}
	\begin{table}[t]
		\begin{center}
			\centering
			\renewcommand\tabcolsep{3pt}
			\begin{tabular}{c | c c | c c}
				\hline
				Method&DS3L&MTCF&Ours (w/o OOD)&Ours\\
				\hline
				Accuracy      & 4.50   & 29.05 & 59.90 & 62.53 \\ \hline
			\end{tabular}
		\end{center}
		\caption{Accuracy($\%$) on TIN 20 using different methods.}
		\label{table:Ours-TIN}
	\end{table}
	\end{comment}
	
	\subsection{Results and analysis}\label{sec_5_2}

	Experiment results of Style Disturbance with PL and DACT are listed in Table \ref{table:MixMatch-vs-UDA-vs-DS3L-c10}, Table \ref{table:PL and MixStyle}, respectively.
	The results show that: OOD samples can stably boost performance, instead of harming it in Table \ref{table:PL_methods};
	With the help of OOD samples, Style Disturbance can help boost model performance on both PL and DACT based SSL methods.

	Experiment results on CIFAR-10 shows great performance of our method, as is listed in Table \ref{table:MixMatch-vs-UDA-vs-DS3L-c10}.
	Compared to previous SOTA methods of MTCF \cite{yu2020multi} and DS3L \cite{guo2020safe}, our method outperforms the others not only on classification performance but also on robustness: previous methods avoid disadvantage of OOD samples by reducing their weight, and degradation of model performance is mitigated notably; 
	however, our method tries to take advantages of OOD samples, and improves model performance by 3.35$\%$ and 0.92$\%$ respectively after using them.

	More experiments and analyses are provided in the supplementary material, such as: 
	(1) Analyzing the relationship between DACT robustness and the distribution gap of ID \& OOD samples, to show that Style Disturbance does reduce the gap;
	(2) Implementation details and performances of OOD detection module to show its effectiveness;
	(3) Ablation studies on Style Disturbance \& DACT to better understand the contribution of each module.

	\section{Conclusion}
	We analyze the robustness of two fundamental SSL methods: PL and DACT, to the more realistic open-set SSL setting. 
	Our observation reveals that:
	(1) DACT is more robust to OOD samples than PL. 
	However, data augmentation for DACT needs to be diverse and carefully searched. 
	(2) DACT on Open-Set SSL has close relationships with multi-view consistency based self-supervised learning in terms of the loss formulation and similar experimental phenomena. 
	(3) OOD samples can be better utilized for PL and DACT by our proposed method Style Disturbance. 
	Experiments on several open-set SSL benchmarks prove that our method achieve better performance than previous SOTA methods.

	\small\bibliography{aaai22}

	%%%%%%%%% BODY TEXT
	\section{More experiments on Pseudo Labeling}
	\subsection{Pure Pseudo Labeling on more datasets}
	We further conduct experiments on Tiny ImageNet (TIN) \cite{deng2009imagenet} and Street View House Numbers (SVHN) \cite{netzer2011reading} with the basic Pseudo Labeling (PL) based Semi-Supervised Learning (SSL) method,  Pseudo-Label \cite{lee2013pseudo}, to verify that PL-based SSL methods are generally not robust on open-set SSL setting.
	
	\noindent\textbf{Experiment setting:} For SVHN, 7.3k samples are split from the original training data as validation dataset; the remaining samples are split into labeled and unlabeled dataset; number of labeled samples is 1k. 10k OOD samples are added into unlabeled dataset from the following 4 datasets for each setting: TIN, LSUN \cite{yu2015lsun}, Gaussian noise (GN) and Uniform Noise (UN).
	
	\noindent\textbf{Implementation detail:} We follow the same setting as is mentioned in the main body of the paper.
	
	As is shown in Table \ref{table:Pseudo_Label on 2 datasets}, Pseudo-Label is also not robust on TIN and SVHN, and it again verifies our point of view.
	
	\subsection{Pseudo Labeling combined with other SSL methods}
	We also analyze the PL module in MixMatch \cite{berthelot2019mixmatch}, an SSL algorithm make up of both PL and DACT methods.
	We compare the model performance before and after removing the PL module to show that PL is not robust to OOD samples.
	
	The PL module of MixMatch is integrated in MixUp \cite{zhang2017mixup}, a regularization method / augmentation policy widely used in the field of deep learning.
	MixMatch first puts pseudo-labels on unlabeled samples, then these pseudo-labeled samples are mixed up with labeled samples $D_L^{ori}$ in both image level and label level, to generate labeled dataset $D_L$ used for training.
	We argue that $D_L$ could suffer from the negative impact caused by pseudo-labeled  OOD samples, so we replace $D_L$ with $D_L^{ori}$ for training. 
	
	The open-set SSL experiment is conducted on CIFAR-10 in the same setting as is mentioned in the DACT section of the main paper.
	The results are listed in Table \ref{table:MixMatch-vs-UDA-c10}.
	We can observe that: though replacing $D_L$ with $D_L^{ori}$ would let model performance drop a little on conventional SSL setting, the performance degradation is mitigated remarkablely (by 1.67\% and 0.94\% respectively) when we do this in open-set SSL setting.
	The phenomena again indicate that PL is generally not robust to OOD samples in open-set SSL setting.

	\section{DACT robustness and distribution gap between labeled $\&$ unlabeled data}
	\subsection{Distribution gap measurement}
	In the field of transfer learning, the distribution change or domain shift caused by many factors (e.g., image quality, modality, style) between the source and target domain can always degrade model performance \cite{wang2018deep, csurka2017domain}.
	Maximum Mean Discrepancy (MMD) \cite{gretton2007kernel} is widely used to measure the distribution gap between domains \cite{long2013transfer, wangimportance, hou2016unsupervised}.
	Our experiment results on open-set SSL also show the variance of model performance in terms of OOD samples.
	Following studies \cite{long2013transfer, wangimportance, hou2016unsupervised} to measure the gap better by balancing both label and structural information, we adopt MMD and class-wise MMD \cite{long2013transfer} together to evaluate the marginal and conditional distribution gap between labeled and unlabeled dataset.
	
	The metric $mmd_{gap}$ is written as: 
	\begin{equation}
		\begin{aligned}
			&mmd_{gap} = \|\frac{1}{|D_L|}\sum\limits_{i=1}^{|D_L|}p_{\phi}(y|x_i)-\frac{1}{|D_U|}\sum\limits_{j=1}^{|D_U|}p_{\phi}(y|x_j)\|_{\mathcal{H}}^2 \\
			&+ \sum\limits_{c=1}^{K}\|\frac{1}{|D_L^{c}|}\sum\limits_{x_i \in D_L^c}p_{\phi}(y|x_i)-\frac{1}{|D_U^c|}\sum\limits_{x_j \in D_U^c}p_{\phi}(y|x_j)\|_{\mathcal{H}}^2
		\end{aligned}
	\end{equation}
	where K is the number of class in $D_L$, $D_L^c$ is the labeled c-th class dataset, $D_U^c$ is the c-th dataset pseudo-labeled via maximal probability, $\phi$ is the model, $x$ is the input image and $\phi(x)$ is a 1xK vector representing the probability of $x$ to each class.
	
	\subsection{Experiment result}
	We use $mmd_{gap}$ on CIFAR-10 with DACT based SSL method, UDA, to observe the change of performance in terms of distribution gap. Model $\phi$ is trained by $D_L$ only. 
	The last columns of Table \ref{table:MixMatch-vs-UDA-c10} show how the distribution gap varies with the OOD samples of unlabeled dataset. 
	We notice that the top-1 accuracy of model trained by UDA rises when $mmd_{gap}$ drops, indicating that the distribution gap also results in the change of performance on open-set SSL setting. 
	In a word, a similar phenomenon appears in both the transfer learning and the open-set SSL setting: 
	the model performs better as the distribution gap between $D_L$ (target domain) and $D_U$ (source domain) gets smaller.
	
	\begin{table}
		\begin{center}
			\centering
			\begin{tabular}{c|c|c|c}
				\hline
				ID dataset              & OOD dataset    & Accuracy  & \makecell[c]{Mean acc\\ change}  \\ \hline
				\multirow{5}{*}{SVHN} & Clean   &  92.76 ± 0.32 & \multirow{5}{*}{\color{red} $\downarrow$ 0.50} \\ 
				{}              & LSUN  & 92.09 ± 0.37        & {} \\ 
				{}              & TIN    & 91.62 ± 0.44   & {} \\ 
				{}              & Gaussian       & 92.65 ± 0.21    & {} \\ 
				{}              & Uniform        & 92.70 ± 0.15   & {} \\ \hline
				\multirow{2}{*}{TIN 20} & Clean & 44.10 ± 0.33 & \multirow{2}{*}{\color{red} $\downarrow$ 0.53} \\ 
				{}              & TIN 180 & 43.57 ± 0.61 & {} \\ \hline
			\end{tabular}
		\end{center}
		\caption{Performance of Pseudo-Label on SVHN and TIN 20.}
		\label{table:Pseudo_Label on 2 datasets}
	\end{table}
	
	\begin{table*}[!t]
		\begin{center}
			\centering
			\begin{tabular}{c|c|c|c|c|c|c}
				\hline
				\multirow{2}{*}{Method} &
				\multicolumn{5}{c|}{OOD dataset} & \multirow{2}{*}{Mean acc change} \\
				\cline{2-6}
				{} & Clean & LSUN & TIN & Gaussian & Uniform & {} \\
				\hline\hline
				
				MixMatch & 90.67 ± 0.29 & 87.03 ± 0.41 & 88.03 ± 0.22 & 84.49 ± 1.06 & 85.71 ± 1.14 & \color{red} $\downarrow$ 4.36 \\
				MixMatch(w/o PL) & 90.08 ± 0.29 & 87.93 ± 0.17 & 88.64 ± 0.20 & 86.09 ± 1.27 & 86.90 ± 0.12 & \color{red} $\downarrow$ 2.69 \\
				UDA & 88.29 ± 0.25 & 88.60 ± 0.22 & 88.86 ± 0.37 & 89.15 ± 0.19 & 89.22 ± 0.25 & \color{blue} $\uparrow$ 0.67 \\
				\hline\hline
				$mmd_{gap}$ & / & 1.71 ± 0.26 & 1.47 ± 0.26 & 1.20 ± 0.20 & 1.19 ± 0.19 & / \\
				\hline
			\end{tabular}\label{MMvsUDA-1k}
			\caption{CIFAR-10 with 1000 labeled and 54000 unlabeled samples}
		\end{center}
		%\end{table*}
		
		%\begin{table*}
		\begin{center}
			\centering
			\begin{tabular}{c|c|c|c|c|c|c}
				\hline
				\multirow{2}{*}{Method} &
				\multicolumn{5}{c|}{OOD dataset} & \multirow{2}{*}{Mean acc change} \\
				\cline{2-6}
				{} & Clean & LSUN & TIN & Gaussian & Uniform & {} \\
				\hline\hline
				
				MixMatch & 93.30 ± 0.10 & 91.18 ± 0.33 & 91.25 ± 0.13 & 90.47 ± 0.38 & 91.51 ± 0.35 & \color{red} $\downarrow$ 2.20 \\
				MixMatch(w/o PL) & 93.11 ± 0.03 & 91.56 ± 0.02 & 91.98 ± 0.09 & 92.09 ± 0.11 & 91.79 ± 0.24 & \color{red} $\downarrow$ 1.26 \\
				UDA & 93.36 ± 0.40 & 93.56 ± 0.04 & 93.65 ± 0.14 & 93.80 ± 0.08 & 93.84 ± 0.47 & \color{blue} $\uparrow$ 0.35 \\
				\hline\hline
				$mmd_{gap}$ & / & 1.85 ± 0.35 & 1.54 ± 0.37 & 1.28 ± 0.32 & 1.26 ± 0.31 & / \\
				\hline
			\end{tabular}
			\caption{CIFAR-10 with 4000 labeled and 51000 unlabeled samples}
		\end{center}
		\caption{Accuracy($\%$) for CIFAR-10 and OOD dataset pairs. Following the setup in \cite{yu2020multi}, we report the averages and the standard deviations of the scores obtained from three trials. "Clean" means unlabeled dataset doesn't contain OOD samples.}
		\label{table:MixMatch-vs-UDA-c10}
	\end{table*}
	
	\section{DACT robustness and diversity of augmentation strategy}
\begin{figure}[htbp]
		\begin{center}
			\centering
			\includegraphics[width=1.1\linewidth]{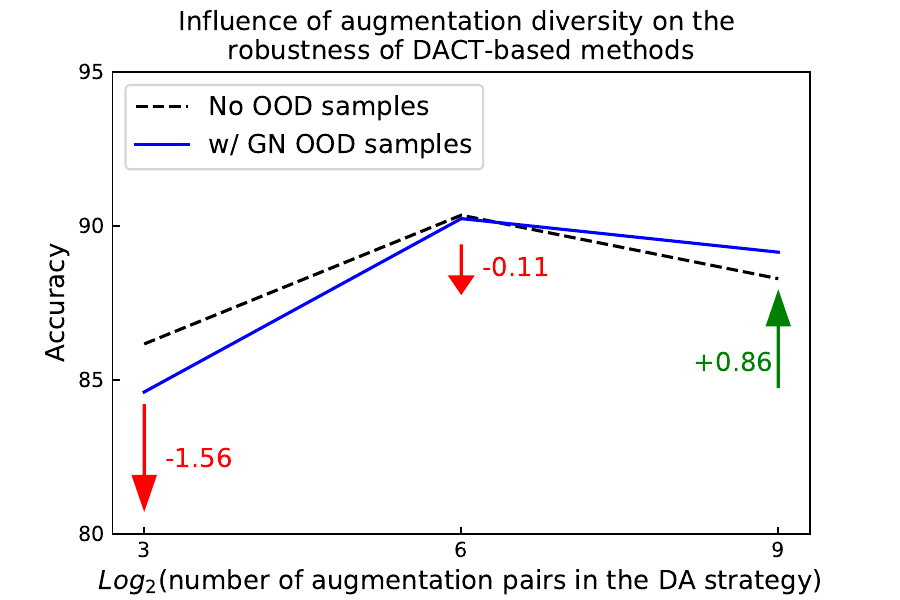}
		\end{center}
		\caption{Change of DACT robustness in terms of DA strategy's diversity. Augmentation pairs ranked in top \{8, 62, 493\} are chosen to form new DA strategy. Experiments are conducted with UDA \& FA for CIFAR-10 on CIFAR-10 with GN samples. When the diversity of DA strategy grows larger, model performs more robustly on OOD samples.}
		\label{fig:robustness and diversity}
	\end{figure}

	To better understand the relationship between DACT robustness on OOD samples and diversity of DA strategies, we vary the diversity to observe the change of robustness. 
	Fast AugoAugment (FA) is a DA strategy composed of augmentation pairs, and we could use the number of pairs to estimate its diversity roughly. 
	The FA augmentation strategy searched for CIFAR-10 is made up of $N_{DA} = 493$ augmentation pairs. 
	We take number of augmentation pairs of this strategy from \{8, 62, 493\} ($N_{DA} / 2^{0} = 493, N_{DA} / 2^{3} = 62, N_{DA} / 2^{6} = 8$) to construct new strategies, so as to vary the diversity of DA strategy.
	We do experiments on CIFAR-10 (1k labeled, 54k unlabeled) with 10k unlabeled GN samples to observe the performance change before and after using OOD samples.
	Figure \ref{fig:robustness and diversity} visualizes the experiment result of the experiment above:
	the model tends to be more robust to OOD samples when the diversity of DA strategy grows larger.
	
	Since AutoAugment (AA) \cite{cubuk2018autoaugment}, Fast AutoAugment (FA) \cite{lim2019fast} and RandAugment (RA) \cite{cubuk2020randaugment} are searched in the same augmentation space, we could use the combinatorial number of all possible augmentation results as a metric to estimate the diversity of these DA strategies roughly.
	The combinatorial number can be calculated easily: 
	every single DA policy used in AA, FA or RA is in the form of "[policy \textbf{N}ame, \textbf{I}ntensity, \textbf{P}robability]", which means each policy is adopted with certain intensity and probability.
	\begin{itemize}
		\item {For AA and FA, the DA strategy is make up of many augmentation pairs, e.g., $[[N_{1st}, I_{1st}, P_{1st}], $ $ [N_{2nd}, I_{2nd}, P_{2nd}]]$. Given the augmentation pair number of DA strategy is $N_{DA}$, then the number of possible augmentation combination is $3N_{DA}$.}
		\item{For RA, one augmentation combination is chosen $M$ augmentation policies from 14 types of DA policy with fixed intensity and probability, so the number of possible augmentation combination is: $\sum_{i=1}^{M} C_{14}^{i}$.  $M$ is set as 2 for RA strategy searched on ImageNet.}
	\end{itemize}
	
	Following the main body of the paper, we compare the metric between the generally worst strategy AA, and the best DA strategies among \{RA, FA\} on CIFAR-10, SVHN and TIN datasets. 
	As is shown in Table \ref{table: Diversity of DA on C10 and SVHN} and Table \ref{table: Diversity of DA on TIN}, the model shows better robustness as the diversity of DA strategies gets larger.
	
	\begin{table}
		\begin{center}
			%\begin{subtable}{5.0pt}
			\begin{tabular}{c|c|c|c}
				\hline
				\multirow{2}{*}{\makecell[c]{Data\\aug}} & \multirow{2}{*}{\makecell[c]{ID\\dataset}} &
				\multicolumn{2}{c}{Metric} \\
				\cline{3-4}
				{} & {} & Mean acc change & Diversity  \\
				\hline\hline
				AA & \multirow{2}{*}{CIFAR-10} & 88.87 $\rightarrow$ 87.26 \color{red} (-1.61)  & 285   \\
				FA & {}  & 88.29 $\rightarrow$ 88.88 \color{blue} (+0.59)  & \color{blue}1479  \\
				\cline{1-4}
				AA & \multirow{2}{*}{SVHN}  & 95.54 $\rightarrow$ 95.66 \color{blue} (+0.12) & 120  \\
				FA & {} & 94.85 $\rightarrow$ 95.16 \color{blue} (+0.31) & \color{blue}1491 \\
				\cline{1-4}
				\hline
			\end{tabular}\label{LOSS C10}
			\caption{Diversity of different DA strategies on SVHN and CIFAR-10 with 1k labeled samples.}
			\label{table: Diversity of DA on C10 and SVHN}
			%\end{subtable}

			\begin{tabular}{c|c|c|c}
				\hline
				\multirow{2}{*}{\makecell[c]{Data\\aug}} & \multirow{2}{*}{\makecell[c]{ID\\dataset}} &
				\multicolumn{2}{c}{Metric} \\
				\cline{3-4}
				{} & {} & Mean acc change & Diversity  \\
				\hline\hline
				AA & \multirow{2}{*}{TIN 20} & 54.55 $\rightarrow$ 51.40 \color{red} (-3.15) & 60   \\
				RA & {} & 51.65 $\rightarrow$ 53.35 \color{blue} (+1.70) & \color{blue} 105  \\
				\cline{1-4}
				\hline
			\end{tabular}\label{LOSS C10}
			\caption{Diversity of different DA strategies on TIN 20 with 1k labeled samples.}
			\label{table: Diversity of DA on TIN}
		\end{center}
	\end{table}
	
	\section{OOD detection}
	\subsection{Design of the module}
	An efficient OOD detection module is important both for robust PL-based SSL methods and better use of OOD samples. 
	The detection module of our DACT \& Style Disturbance is a simplified version of previous method \cite{yu2020multi}:
	The two projection heads designed for K-class image classification and OOD sample detection are merged into one (K+1)-class head, and the (K+1)-th class denotes the probability of samples to be OOD. 
	All unlabeled samples are regarded as OOD samples at the beginning of training. 
	To prevent unlabeled samples from being split into ID samples too early, the prediction of original image $p_{\phi}(y|x)$ in consistency training loss is replaced with $p_{\phi}^{t}(y|x)$, the weighted sum of model's previous predictions $p_{\phi}^{t-1}(y|x)$ and current prediction $p_{\phi}(y|x)$: 
	\begin{equation}
		p_{\phi}^{t}(y|x) = \alpha p_{\phi}^{t-1}(y|x) + (1-\alpha) p_{\phi}(y|x)
	\end{equation}
	where the momentum hyper-parameter $\alpha \in [0, 1]$ ($\alpha=0.8$ in experiment). 
	Motivated by momentum update in MoCo \cite{he2020momentum}, previous predictions of $D_U$ are stored in a memory bank. 
	Unlabeled samples are regarded as OOD ones if the $(K+1)$-th probability of output is the largest after half of total epochs.

	\subsection{Performance of the module}
	We compare our splitting method to existing OOD detection method \cite{yu2020multi} to verify its effectiveness.
	For simplicity, we only evaluate OOD detection performance on CIFAR-10 setting with 1,000 labeled samples; 
	Since more labeled ID samples will surely make it easier to filter out OOD samples, we do not conduct the CIFAR-10 setting with 4,000 labeled samples here.
	As is shown in Table \ref{table:ood_det}, as a simplified version of MTCF \cite{yu2020multi}, our splitting module performs comparably well.
	
	\section{More experiments on the use of OOD samples}
	\subsection{Ablation studies on Style Disturbance \& DACT}
	To verify the generalization of methods, we turn to TIN with a larger backbone ResNet-50. 
	Using the official implementation of DS3L $\&$ MW-Net \cite{shu2019meta} (backbone of DS3L) and MTCF, our experiments show that both methods have unsatisfactory performance. 
	Besides, hardly can I tune hyper-parameters because both methods are very time-consuming and memory-unfriendly, as is shown in Table \ref{table:MTCF-DS3L-Ours-TIN}. 
	Also we could not find any references help guide the hyper-parameter tuning procedure for either method on ImageNet or Tiny ImageNet. 
	Consequently, we only make several trials for each method, and report the best result of them. 
	In contrast to the above two methods, ours is much faster and far more robust on OOD samples. 
	As is shown in the 5-th row of Table \ref{table:Ours-TIN}, our method enhances model performance by 2.63$\%$.
	
	The quickly advancing field of Self-Supervised Learning \cite{he2020momentum, chen2020improved, chen2020simple, chen2020big, caron2020unsupervised} also motivates us to make better use of OOD samples for better pretrained models. 
	We simply choose MoCo \cite{he2020momentum}, a GPU-friendly method to pretrain the model with both $D_L$ and $D_U$. 
	The pretrained model is then used to initialize the network for subsequent tasks.
	
	We perform an ablation study on TIN to better understand how each module works. 
	We analyze the effect of components in our method and find that each module has an orthogonal contribution to the overall improvements, as is summarized in Table \ref{table:Ours-TIN}. 
	We observe that: 
	(1) Adding 90,000 OOD samples to $D_U$ directly could bring about 1$\%$ improvement, and it again verifies the robustness of DACT; 
	(2) Module of style disturbance boosts the performance more than 1.5$\%$; Apart from splitting module, it still contributes about 1$\%$ to improvement;
	Since the splitting module performs in a similar way as another widely-used SSL method Temporal Ensembling \cite{laine2016temporal}, it brings roughly 0.5$\%$ improvement as well; 
	(3) The pretraining module also shows the value of OOD dataset by enhancing performance for about 1.6$\%$; 
	(4) The combination of all components improves totally 5.8$\%$ and shows great advantage on other open-set SSL methods. 
	
	\subsection{Why using self-supervised pretraining?}
	Self-supervised learning shows great performance to use OOD samples to improve model performance on downstream ID tasks.
	As is discussed in the main body of our paper, DACT on open-set SSL could gain from OOD samples in a similar way as multi-view consistency based self-supervised learning; 
	however, two learning paradigms are quite different from each other in some dimension. 
	For example, the loss function of consistency loss for DACT, and the InfoNCE \cite{oord2018representation} loss for self-supervised contrastive learning are quite different: 
	the latter aims to pull features of every two samples in the dataset away, while the former tends to draw features from the same class closer. 
	Besides, the use of unlabeled ID data in open-set SSL always show positive influence; 
	in self-supervised learning, however, sometimes taking limited downstream ID data into self-supervised representation learning phase would hurt the final performance \cite{teng2021can}.
	Thus the two paradigms also utilize OOD samples differently to some extent, which means they can benefit model performance orthogonally. Experiments in Table \ref{table:Ours-TIN}  and experiments of OpenCoS \cite{Jong2020opencos} also prove this.

	\subsection{Implementation of self-supervision module}
	To obtain better visual representations with OOD samples, Moco v1 \cite{he2020momentum} is adopted as the self-supervised pretraining module of our method. 
	We implement the pretraining module based on the official implementation of MoCo v1. 
	We take the default hyper-parameter to pretrain the model for 200 epochs, and the pretraining procedure takes about 15 hours on TIN \cite{deng2009imagenet}. 
	
	\begin{table}
		\begin{center}
			\centering
			\begin{tabular}{c c c c|c}
				\hline
				w/ OOD & split & style trans & pretrain & top-1 acc \\ \hline
				-          & -          & -          & -          & 59.90±0.54 \\
				\checkmark & -          & -          & -          & 60.95±0.55 \\
				\checkmark & \checkmark & -          & -          & 61.50±0.12 \\
				\checkmark & \checkmark & \checkmark & -          & 62.53±0.24 \\
				\checkmark & -          & -          & \checkmark & 62.58±0.79 \\
				\checkmark & \checkmark & \checkmark & \checkmark & \textbf{65.70±0.16} \\ \hline
			\end{tabular}
		\end{center}
		\caption{Ablation studies on Tiny ImageNet.}
		\label{table:Ours-TIN}
	\end{table}
	
	\begin{table}
		\begin{center}
			\centering
			\begin{tabular}{c|c|c|c}
				\hline
				Method                  & Time/trial           & Device num  & Top-1 acc  \\ \hline
				DS3L & \textgreater 1 week  & 8           & 4.50        \\
				MTCF & \textgreater 2 weeks & 8           & 29.05       \\
				Ours                    & \textless 40 hours        & 2           & \textbf{65.70±0.16} \\ \hline
			\end{tabular}
		\end{center}
		\caption{Comparison of different methods on Tiny ImageNet. The device we used is NVIDIA Tesla V100.}
		\label{table:MTCF-DS3L-Ours-TIN}
	\end{table}
	
	\begin{figure}[htbp]
		\begin{center}
			\centering
			\includegraphics[width=1.0\linewidth]{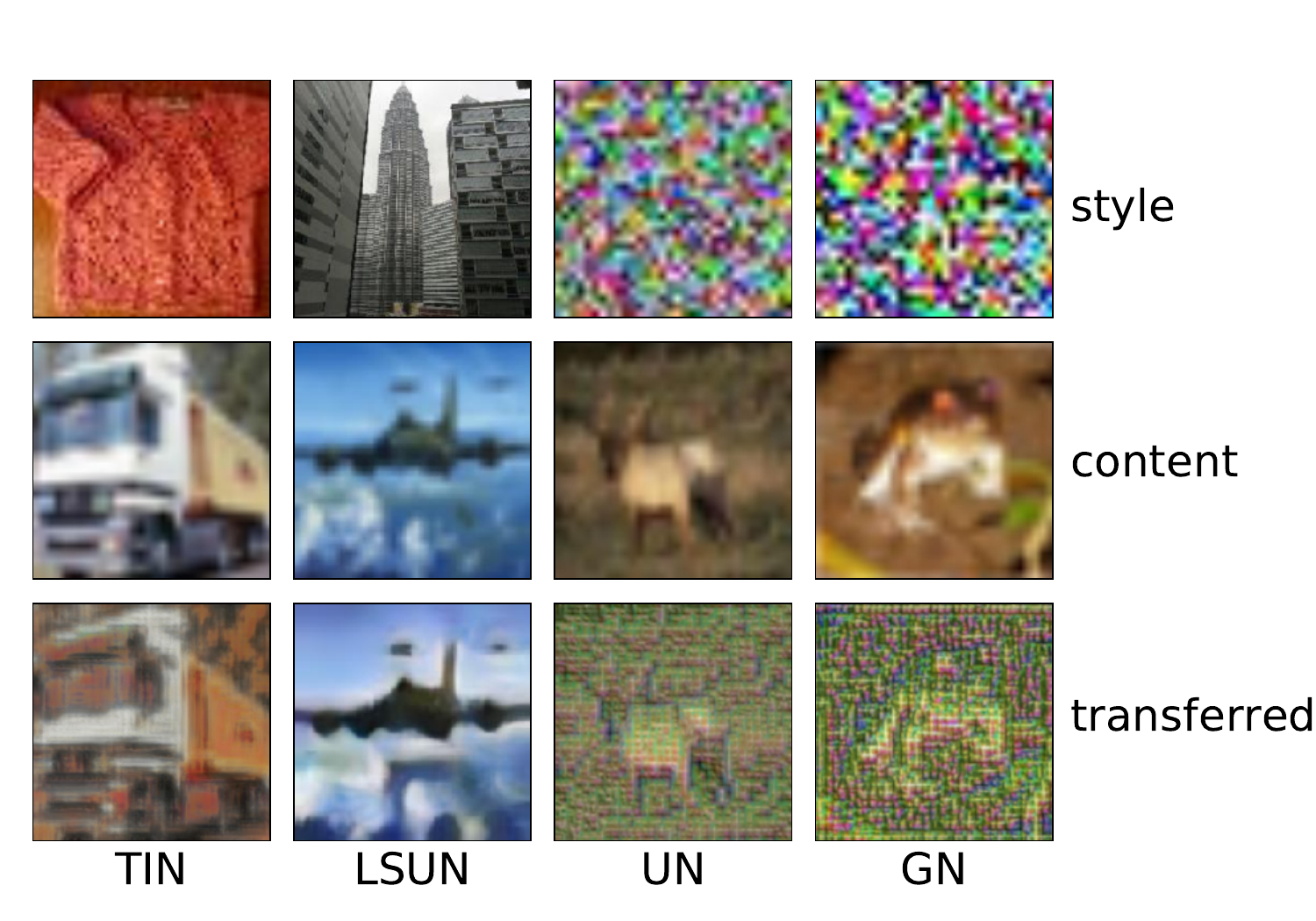}
		\end{center}
		\caption{Style-disturbed images in CIFAR-10 setting. Styles of four OOD datasets, Tiny ImageNet (TIN), Large-scale Scene Understanding dataset (LSUN), Uniform Noise (UN) and Gaussian Noise (GN), are transferred to ID images.}
		\label{fig:st_cifar10}
	\end{figure}
	
	\begin{figure}[htbp]
		\begin{center}
			\centering
			\includegraphics[width=1.0\linewidth]{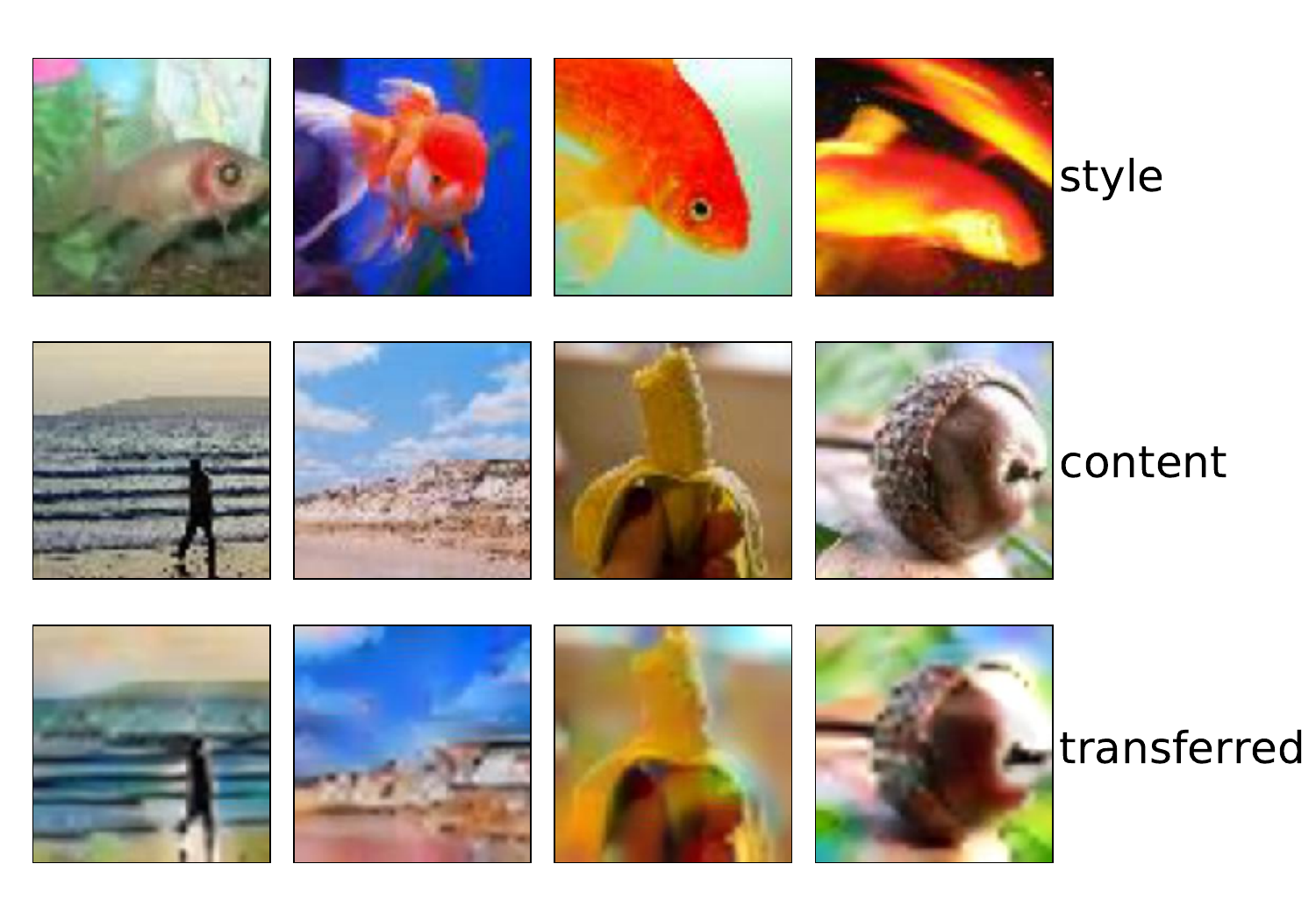}
		\end{center}
		\caption{Style-disturbed images in TIN setting.}
		\label{fig:st_TIN_99k}
	\end{figure}
	
	\begin{table*}[h]
		\begin{center}
			\centering
			\begin{tabular}{|c|c c|c c|c c|c c|}
				\hline
				\multirow{3}{*}{Method} &
				\multicolumn{8}{c|}{OOD dataset} \\
				\cline{2-9}
				{} & \multicolumn{2}{c|}{TIN} & \multicolumn{2}{c|}{LSUN} & \multicolumn{2}{c|}{UN} & \multicolumn{2}{c|}{GN} \\
				\cline{2-9}
				{} & Precision & Recall & Precision & Recall & Precision & Recall & Precision & Recall \\
				\hline
				MTCF & 97.11 & 99.48 & 98.95 & 99.95 & 100 & 100 & 100 & 100 \\
				Ours  & 97.22 & 99.65 & 98.20 & 99.34 & 100 & 100 & 100 & 100 \\
				\hline
			\end{tabular}
		\end{center}
		\caption{The comparison of precision and recall($\%$) on OOD detection tasks in CIFAR-10 setting with 1,000 labeled samples.}
		\label{table:ood_det}
	\end{table*}
	
	%-------------------------------------------------------------------------
	\section{Visualization of style-disturbed images}
	Style-disturbed images of both CIFAR-10 \cite{krizhevsky2009learning} and TIN setting are shown in Figure \ref{fig:st_cifar10} and Figure \ref{fig:st_TIN_99k} respectively.
	From these images we can observe that the model transfers the style of OOD images to ID images slightly, while the content of image is preserved.
	
	\section{Implementation of other methods}
	\noindent\textbf{MTCF} \cite{yu2020multi}: Our implementation is based on the official PyTorch implementation code.
	For CIFAR-10 setting, we use the experiment result reported in the paper.
	For TIN setting, we simply replace the dataset, and try to take the default hyper-parameter to train the model.
	Since it takes too much time to train for a single trial, we just run 2 trials and report the higher result.
	
	\noindent\textbf{DS3L} \cite{guo2020safe}: Our implementation is based on the official PyTorch implementation code.
	For CIFAR-10 setting, we modify the ID and the OOD dataset, then directly take the default hyper-parameter to train the model.
	For TIN setting, we meet a few challenges during training:
	\begin{itemize}
		\item {
			In the data pre-processing process, we find it impossible to apply either global contrast normalization or ZCA-normalization to dataset because of our server's limited capacity of memory (about 375G). 
			So we have to remove the two steps, and replace them with the common normalization of ImageNet \cite{deng2009imagenet} as is done in our method.}
		\item {
			Since DS3L takes the MW-Net \cite{shu2019meta} as backbone, we transfer the original ResNet-50 network into the format of MW-Net for training. 
			We also take the default hyper-parameter for training. As it takes too much time to train for a single run, we just run 1 trial and report the result.}
	\end{itemize}
	
	From our experiment, we find that the above two methods do not have robust hyper-parameters; 
	Besides, both of the official implementation require a long time to train on a larger dataset. 
	The two problems might be a large obstacle when applying these methods to realistic settings.

\end{document}